\documentclass[11pt]{article}
\usepackage[utf8]{inputenc}
\usepackage{times}
\usepackage[margin=1in]{geometry}
 
\usepackage{mathtools}
\usepackage{xcolor}
\usepackage{hyperref}
\definecolor{bluegray}{rgb}{0.3, 0.5, 0.7}
\hypersetup{
    colorlinks=true,
    linkcolor=bluegray,
    urlcolor=bluegray,
    citecolor=bluegray
}
\usepackage{cleveref}
\usepackage{subfig}
\usepackage{booktabs}
\usepackage{amssymb}
\usepackage{natbib}

\newcommand{\DGL}{DGL}
\setcitestyle{numbers}

\begin{document}
\noindent {\color{blue}{The published version of this article can be found at \url{https://doi.org/10.1073/pnas.2400917121}. Please cite the published version as: \\

\noindent Hu, J., Mahowald, K., Lupyan, G., Ivanova, A., \& Levy, R. (2024). Language models align with human judgments on key grammatical constructions. \textit{Proceedings of the National Academy of Sciences}, 121(36), e2400917121. https://doi.org/10.1073/pnas.2400917121}}

\bigskip

\vspace*{1cm}

\begin{center}
    \Large{Language models align with human judgments on key grammatical constructions} 
    \bigskip
    \bigskip
    \bigskip
    
    \large{Jennifer Hu\textsuperscript{1}, Kyle Mahowald\textsuperscript{2}, Gary Lupyan\textsuperscript{3}, Anna Ivanova\textsuperscript{4}, Roger Levy\textsuperscript{5}}

    \bigskip
    
    \normalsize
    \textsuperscript{1}Kempner Institute for the Study of Natural and Artificial Intelligence, Harvard University \\
    \textsuperscript{2}Department of Linguistics, The University of Texas at Austin\\
    \textsuperscript{3}Department of Psychology, University of Wisconsin-Madison \\
    \textsuperscript{4}School of Psychology, Georgia Tech\\
    \textsuperscript{5}Department of Brain and Cognitive Sciences, Massachusetts Institute of Technology
\end{center}

\vspace{3cm}

\noindent Author contributions:
\begin{table}[ht]
    \centering
    \scriptsize
    \begin{tabular}{cccccc} \toprule
        Author & Designed research & Performed research & Contributed new reagents/analytic tools & Analyzed data & Wrote the paper \\ \midrule
        JH & \checkmark & \checkmark & & \checkmark & \checkmark \\
        KM & \checkmark &  & & \checkmark & \checkmark \\
        GL & \checkmark & \checkmark & & \checkmark & \checkmark \\
        AI & \checkmark & & & \checkmark & \\
        RL & \checkmark & & & \checkmark & \checkmark \\ \bottomrule
    \end{tabular}
\end{table}

\vspace{3cm}

\noindent Corresponding authors:\\
Jennifer Hu (jenniferhu@fas.harvard.edu) or Roger Levy (rplevy@mit.edu)

\thispagestyle{empty}
\newpage

\begin{center}
    \large \textbf{Language models align with human judgments on key grammatical constructions}
\end{center}
Do large language models (LLMs) make human-like linguistic generalizations? \citet{dentella_systematic_2023} (DGL) prompt several LLMs (``Is the following sentence grammatically correct in English?'') to elicit grammaticality judgments of 80 English sentences, concluding that LLMs demonstrate a ``yes-response bias'' and a ``failure to distinguish grammatical from ungrammatical sentences''. 
We re-evaluate LLM performance using well-established practices and find that \DGL{}'s data in fact provide evidence for how well LLMs capture human linguistic judgments.\footnote{We release our code and data at the following link: \url{https://github.com/jennhu/response-to-DGL}}

\begin{figure}[p]
    \centering
    \subfloat[\label{fig:minimal-pairs-accuracy}]{
        \includegraphics[width=0.9\linewidth]{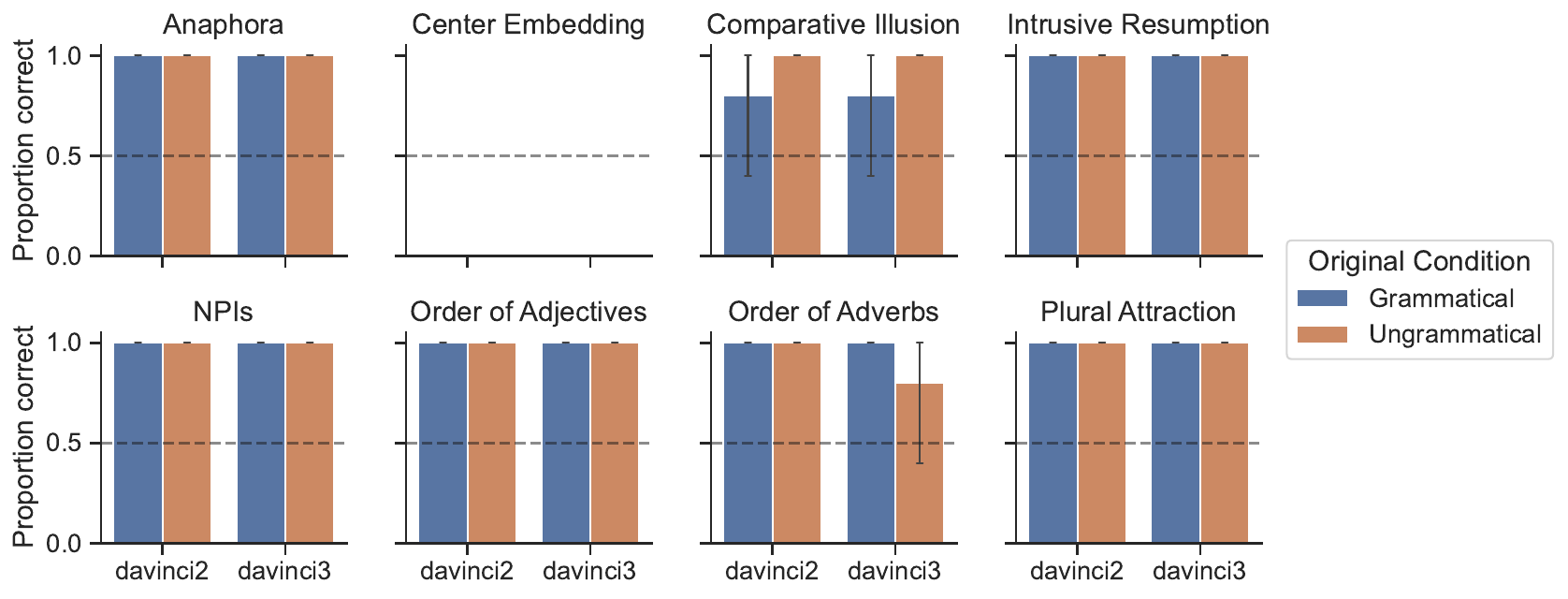}
    }\\
    \subfloat[\label{fig:minimal-pairs-surprisal-difference}]{
        \includegraphics[width=0.9\linewidth]{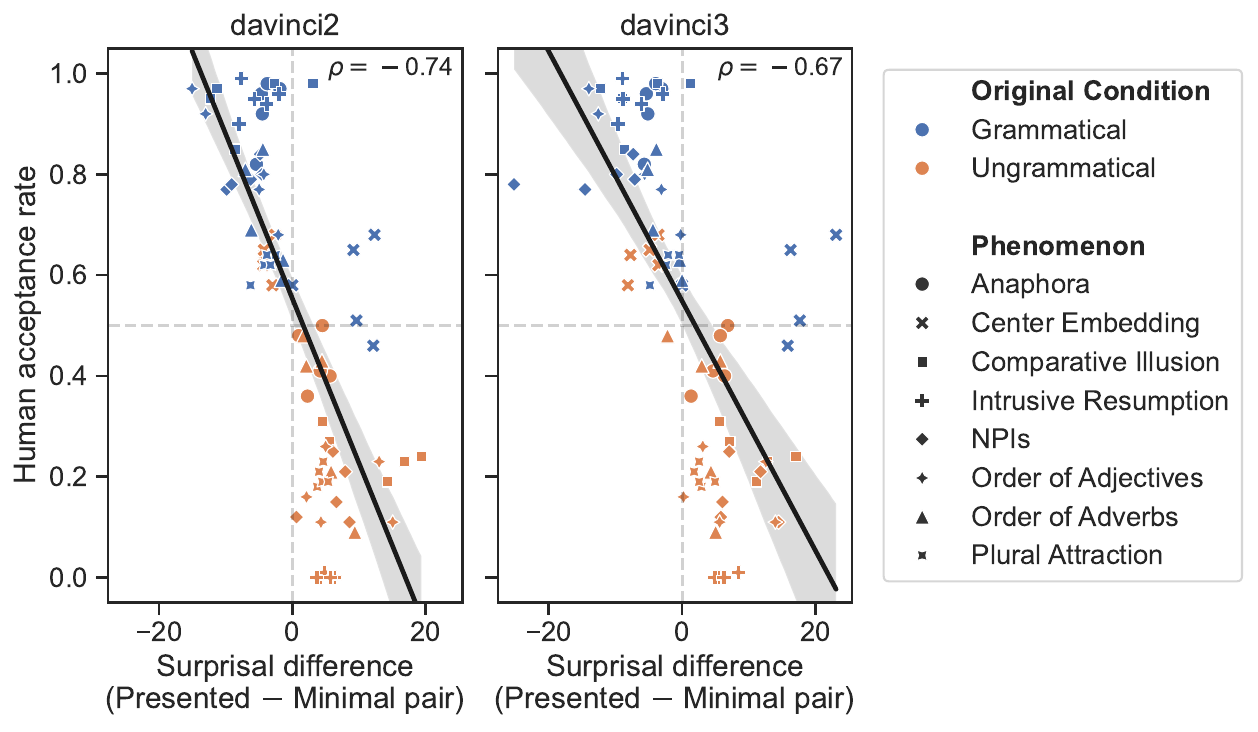}
    }
    \caption{(a) Accuracy scores achieved by models on a version of DGL's original materials with minimal pairs. For each phenomenon, accuracy is computed as the proportion of items in that phenomenon where the model assigns higher probability to the grammatical version of that item (minimal pair) than the ungrammatical version. (b) x-axis: Difference in sum surprisal (negative log probability) between the sentence presented to humans in \DGL's experiments versus its counterpart in the minimal pair. y-axis: Human acceptance rate (proportion judged as grammatical) for the presented sentence in each minimal pair. Each point represents a minimal pair test item.}
    \label{fig:minimal-pairs}
\end{figure}

The ability to produce well-formed sentences does not necessarily require being able to articulate the underlying grammatical rules. This distinction has been long noted in linguistics \citep[e.g.,][]{birdsong_metalinguistic_1989,han_grammaticality_2000,chomsky_knowledge_1986}, but is blurred by DGL: their task tests not only LLMs' grammatical competence, but also whether models know what ``grammatically correct'' means. 
To remedy this, we follow standard methods (not discussed by \DGL) of evaluating LLMs' linguistic knowledge  
\citep[e.g.,][]{marvin-linzen:2018-targeted,warstadt_blimp_2020,hu_systematic_2020}. 
Rather than relying on models' metalinguistic skills, a method that systematically underestimates LLM generalization capabilities \citep{hu_prompting_2023}, we directly measure the probabilities models assign to strings \cite{clark-lappin:2010linguistic-nativism}. For each sentence in DGL's materials, we constructed a lexically matched counterpart differing only in the targeted grammatical feature. This controlled manipulation isolates grammatical differences, so a model that has learned the correct generalizations should assign higher probability to the grammatical sentence in each minimal pair. Minimal-pair analysis reveals at- or near-ceiling performance except on Center Embedding (\Cref{fig:minimal-pairs-accuracy}), for which humans are also below chance (47.1\% accuracy). Furthermore, minimal-pair surprisal (negative log-probability) differences predict item-level variation in human responses: the less surprising a sentence relative to its minimal pair, the more likely humans are to judge it as grammatical (\Cref{fig:minimal-pairs-surprisal-difference}; davinci2: Pearson $\rho=-$0.74; davinci3: Pearson $\rho=-$0.67).

Moreover, although \DGL{} argue that human-judgment inaccuracies reflect ``performance factors'', their data reveal systematic variation in human acceptability judgments (Figure 2a). For instance, the Anaphora phenomenon shows two groups of participants: one whose judgments conform to \DGL's labels (bottom right cluster), and one judging all sentences as grammatical (upper right cluster). 
\DGL's logic would imply that only these latter participants suffer performance constraints. Genuine variability in acceptability is a better explanation and is consistent with a wide literature in linguistics \citep{bresnan2007predicting}. For example, the Anaphora sentences that \DGL{} label as ungrammatical use the word ``themselves'' as a singular pronoun, which may be perfectly acceptable to some speakers. Similarly, many participants (43\%) judge Order of Adverbs sentences like ``Gary still perhaps drives to work'' as grammatical, even though \DGL{} code it as ungrammatical.

Finally, \DGL{}'s task differed subtly for models and humans: models were prompted for open-ended responses (which were subsequently coded as correct/incorrect by \DGL{}), whereas humans had to provide a binary judgment by pressing one of two keys. We re-evaluated davinci2, davinci3, GPT-3.5 Turbo, and GPT-4 using the exact prompt seen by humans (Figure 2b). The ``yes''-bias reported by \DGL{} disappears for all models except davinci2. While davinci2 and davinci3 still perform near chance, GPT-3.5 Turbo and GPT-4 \emph{outperform} humans according to \DGL's normative grammaticality coding. Overall, we conclude that LLMs show strong and human-like grammatical generalization capabilities.

\begin{figure}[p]
    \centering
    \subfloat[\label{fig:fig2-acceptance-rates}]{
        \includegraphics[width=\linewidth]{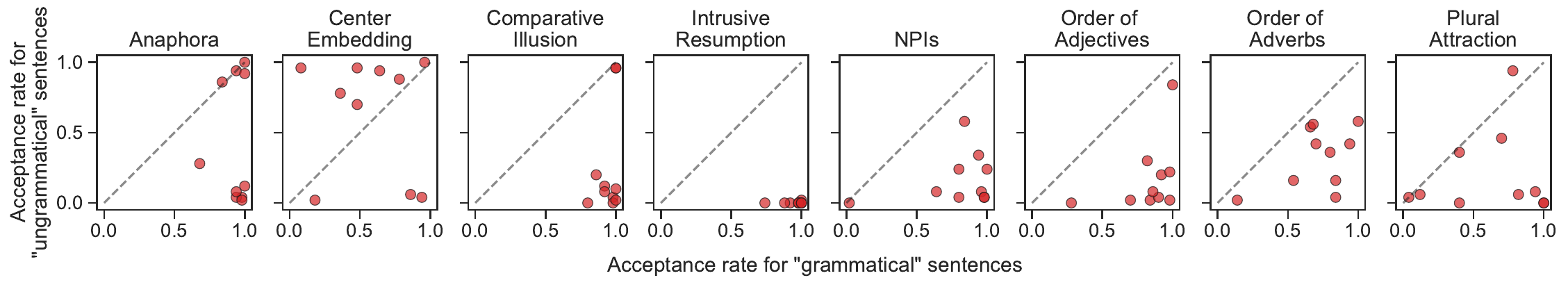}
    }\\
    \subfloat[\label{fig:fig2-confusion-matrices}]{
        \includegraphics[width=\linewidth]{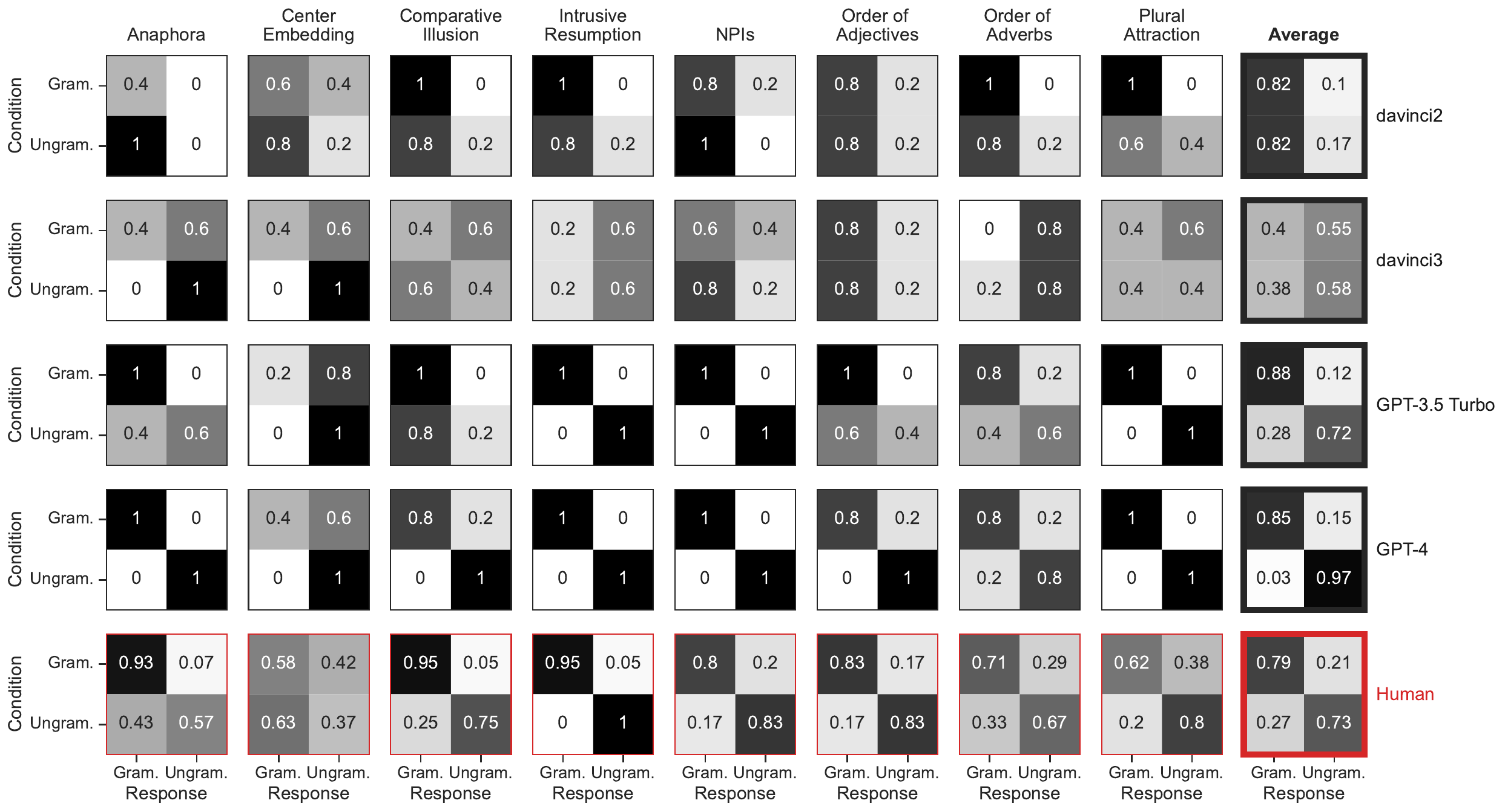}
    }
    \caption{(a) Participant-specific acceptance rates (i.e., rate of judging as grammatical) for sentences that \DGL{} label as ``grammatical'' (x-axis) versus ``ungrammatical'' (y-axis). If participants' responses perfectly reflected \DGL's normative coding, then all participants would be in the bottom right corner (as exemplified by Intrusive Resumption). (b) Confusion matrices achieved by models and humans on each phenomenon, when evaluating models using the same prompt that was seen by humans (``Is the following sentence grammatically correct in English? [SENTENCE] Respond with C if it is correct, and N if it is not correct.''). ``Gram.'' = grammatical, and ``Ungram.'' = ungrammatical. A small fraction of davinci2 and davinci3's responses (4\%) were not codeable as corresponding to ``C'' or ``N'', resulting in missing data.}
    \label{fig:fig2}
\end{figure}

\newpage
\bibliography{references}

\end{document}